\documentclass[letterpaper]{article}
\usepackage{aaai18}
\usepackage{times}
\usepackage{helvet}
\usepackage{courier}
\usepackage{latexsym}

\usepackage{url}

\usepackage{amsfonts,amsmath}
\usepackage{graphicx}
\usepackage{mathtools}
\usepackage{tikz}
\usepackage{url}
\usepackage{multirow}
\usepackage{hhline}

\usepackage[outercaption]{sidecap}    

\usetikzlibrary{arrows,angles,quotes,external}

\newcommand{\inx}{\mathcal{X}}
\newcommand{\outy}{\mathcal{Y}}
\newcommand{\phrasetable}{\mathcal{P}}

\newcommand{\ignore}[1]{}

\newcommand{\decoder}{\mathrm{dec}}

\title{Canonical Correlation Inference for Mapping Abstract Scenes to Text}

\author{
Nikos Papasarantopoulos  \\ School of Informatics \\ University of Edinburgh \\ \texttt{nikos.papasa@ed.ac.uk} \\
\And Helen Jiang \\ Department of Computer Science \\ Stanford University \\ \texttt{helennn@stanford.edu} \\
\And Shay B. Cohen \\ School of Informatics \\ University of Edinburgh \\ \texttt{scohen@inf.ed.ac.uk} \\
}

\newcommand{\newcite}[1]{\citeauthor{#1}~\shortcite{#1}}

\mathtoolsset{showonlyrefs,showmanualtags}

\begin{document}

\maketitle

\begin{abstract}
We describe a technique for structured prediction, based on canonical correlation
analysis. Our learning algorithm finds two projections for the input and the output spaces
that aim at projecting a given input and its correct output into points close to each
other. We demonstrate our technique on a language-vision problem, namely the problem of
giving a textual description to an ``abstract scene.''
\end{abstract}

\section{Introduction}

Canonical correlation analysis (CCA) is a method to reduce the dimensionality of multiview
data, introduced by \newcite{hotelling1935canonical}. It takes two random vectors and computes
their corresponding empirical cross-covariance matrix. It then applies singular value decomposition (SVD)
on this matrix to get linear projections of the random vectors that have maximal correlation.


In this paper, we investigate the idea of using CCA for a full-fledged structured prediction problem.
More specifically, we suggest a method in which we take a structured prediction problem training set,
and then project both the inputs and the outputs to low-dimensional space.
The projection ensures that inputs and outputs that correspond to each other are projected to close
points in low-dimensional space. Decoding happens in the low-dimensional space. As such, our training algorithm builds on previous work by \newcite{udupa2010transliteration} and \newcite{jagarlamudi2012regularized} who
used CCA for transliteration.

Our approach of canonical correlation inference is simple to implement and does not require complex engineering
tailored to the task.
It mainly needs two feature functions, one for the input values and one for the output values and does not require features combining the two.
We also propose a simple decoding algorithm when the output space is text.

We test our learning algorithm on the domain of language and vision. We use the
{\em abstract scene} dataset of \newcite{zitnick2013bringing}, with the goal of mapping images (in the form
of clipart abstract scenes) to their corresponding image descriptions. This problem has a strong relationship
to recent work in language and vision that has used neural networks or other computer vision techniques to solve a
similar problem for real images (Section~\ref{section:related}).
Our work is most closely related to the work by \newcite{ortiz2015learning} who used phrase-based machine
translation to translate images to corresponding descriptions.

\section{Background and Notation}

We give in this section some background information about CCA and the problem which we aim to solve with it.

\subsection{Canonical Correlation Analysis}

Canonical correlation analysis (CCA; Hotelling, 1935\nocite{hotelling1935canonical}) is a technique
for multiview dimensionality reduction, related to co-training \cite{blum1998combining}.
CCA assumes that there are two views for a given set of data, where these two views are represented
by two random vectors $X \in \mathbb{R}^d$ and $Y \in \mathbb{R}^{d'}$.

The procedure that CCA
follows finds a projection of the two views in a shared space of dimension $m$, such
that the correlation between the two views is maximized at each
coordinate, and there is minimal redundancy between the coordinates of
each view.  This means that CCA solves the following sequence of
optimization problems for $j \in \{1 ,\ldots, m \}$ where $a_j \in \mathbb{R}^{1
  \times d}$ and $b_j \in \mathbb{R}^{1 \times d'}$:
\begin{align}
\arg\max_{a_j,b_j} &  & \mathrm{corr}(a_j X^{\top}, b_j Y^{\top}) \\
\mbox{such that} & & \mathrm{corr}(a_j X^{\top}, a_k X^{\top}) = 0, & & k < j \\
		 &  & \mathrm{corr}(b_j Y^{\top}, b_k Y^{\top}) = 0, & & k < j
\end{align}
\noindent where $\mathrm{corr}$ is a function that accepts two vectors
and returns the Pearson correlation between the pairwise elements of
the two vectors. The problem of CCA can be solved by applying singular value decomposition
(SVD) on a cross-covariance matrix between the two random vectors $X$
and $Y$, normalized by the covariance matrices of $X$ and $Y$.

More specifically, CCA is solved by applying thin singular value decomposition (SVD) on the
empirical version of the following matrix:

\begin{equation}
(E[XX^{\top}])^{-\frac{1}{2}} E[XY^{\top}](E[YY^{\top}])^{-\frac{1}{2}} \approx U \Sigma V^{\top},
\end{equation}

\noindent where $E[\cdot]$ is the expectation operator and $\Sigma$ is a diagonal
matrix of size $m \times m$ for some small $m$. Since this version
of CCA requires inverting matrices of potentially large dimension ($d \times d$
and $d' \times d'$), it is often the case that only the diagonal elements
from these matrices are used, as we see in Section~\ref{section:cca-infer}.


CCA and its variants have been used in various contexts in NLP. They were
used to derive word embeddings \cite{dhillon2015eigenwords},
derive multilingual embeddings \cite{faruqui2014improving,lu2015deep},
build bilingual lexicons \cite{haghighi2008learning}, encode prior knowledge
into embeddings \cite{osborne-16}, semantically analyze text \cite{vinokourov2002inferring}
and reduce the dimensions of text with many views \cite{rastogi2015multiview}.
CCA is also an important sub-routine in the family of spectral algorithms for estimating
structured models such as latent-variable PCFGs and HMMs \cite{cohen-12,stratos-16} or finding word
clusters \cite{stratos2014spectral}.
Variants of it have been developed, such as DeepCCA \cite{andrew2013deep}.


\subsection{Describing Images}
\label{section:related}

Image description, the task of assigning textual descriptions to images, is a problem that has been thoroughly studied in various setups and variances.
Usually, proposed methods treat images as sets of objects identified in them (bags of regions),
however there has been work that uses some kind of structural image cues or relations. An excellent example of such cues are visual dependency representations \cite{elliott2013image},
which can be used to outline what can be described as the visual counterpart of dependency trees.

Common is also the idea of solving a related but slightly different problem, the one of matching sentences to images using existing descriptions.
The core of those approaches is an information retrieval task, where for every novel image, a set of similar images is retrieved and generation proceeds 
using the descriptions of those images. Search queries are posed against a visual space \cite{ordonez2011im2text,mason2014nonparametric} or a multimodal space, where 
images and descriptions have been projected \cite{farhadi2010every,hodosh2013framing}. Instead of whole sentences, phrases from existing human generated descriptions 
have also been used \cite{kuznetsova2012collective}.

Approaches to image description generation have for a long time relied on a set of predefined sentence templates 
\cite{kulkarni2011baby,elliott2013image,yang2011corpus} or used syntactic trees \cite{mitchell2012midge}, while more recently, 
methods that use neural models \cite{kiros2014multimodal,vinyals2015show} have appeared, that avoid the use of any kind of predefined pattern. 
Approaches like the latter follow the paradigm of tackling the problem as an end-to-end optimization problem.
\newcite{ortiz2015learning} describe a two-step process: a content selection phase, where the objects that need to 
be described are picked, and then the text realization, where the description is generated 
by employing a statistical machine translation (MT) system.

While computer vision advances have given an unprecedented potential to image description generation, vision performance affects
the generation process, as those two problems are commonly solved together in a pipeline or a joint fashion. 
To countermeasure that, \newcite{zitnick2013bringing} introduced the notion of ``abstract scenes'', that is abstract images generated 
by stitching together clipart images. Their intuition is that working on abstract scenes can allow for a more clean and isolated evaluation
of caption generators and also lead to relatively easy construction of datasets of images with semantically similar content.	
An example of such dataset is the Abstract Scenes Dataset.\footnote{\url{https://vision.ece.vt.edu/clipart/}}
This dataset has been used for description generation \cite{ortiz2015learning}, sentence-to-scene generation \cite{zitnick2013learning} and 
object dynamics prediction \cite{fouhey2014predicting} so far.

\section{Learning and Decoding}
\label{section:cca-infer}

We now describe the learning algorithm, based on CCA, and the corresponding
decoding algorithm when the output space is text.

\subsection{Learning Based on Canonical Correlation Analysis}

We assume two structured spaces, an input space $\inx$ and an output space $\outy$.
As usual in the supervised setting, we are given a set of instances $(x_1,y_1),\ldots,(x_n,y_n) \in \inx \times \outy$,
and the goal is to learn a decoder $\decoder \colon \inx \rightarrow \outy$ such that $\decoder(x)$ is the ``correct''
output as learned based on the training examples.

The basic idea in our learning procedure is to learn two projection functions $u \colon \inx \rightarrow \mathbb{R}^m$
and $v \colon \outy \rightarrow \mathbb{R}^m$ for some low-dimensional $m$ (relatively to $d$ and $d'$). In addition, we assume the existence of
a similarity measure $\rho \colon \mathbb{R}^m \times \mathbb{R}^m \rightarrow \mathbb{R}$ such that for any $x$ and $y$,
the better $y$ ``matches'' the $x$ according to the training data, the larger $\rho(u(x),v(y))$ is.
The decoder $\decoder(x)$ is then defined as:

\begin{figure}[t!]
{\small
\framebox{\parbox{3in}{

{\bf Inputs:} Set of examples $(x_i,y_i) \in \inx \times \outy$ for $i \in \{ 1, \ldots, n\}$. An integer $m$. Two feature functions $\phi(x)$
and $\psi(y)$.

{\bf Data structures:}

Projection matrices $U$ and $V$.

{\bf Algorithm:}

(Cross-covariance estimation)

\begin{itemize}

\item Calculate $\Omega \in \mathbb{R}^{d \times d'}$

\begin{equation}
\Omega_{ij} = \sum_{k=1}^n [\phi(x_k)]_i [\psi(y_k)]_j
\end{equation}

\item Calculate $D_1 \in \mathbb{R}^{d \times d}$ such that $(D_1)_{ij}=0$ for $i \neq j$ and

\begin{equation}
(D_1)_{ii} = \sum_{k=1}^n [\phi(x_k)]_i [\phi(x_k)]_i
\end{equation}

\item Calculate $D_2 \in \mathbb{R}^{d' \times d'}$ such that $(D_2)_{ij}=0$ for $i \neq j$ and

\begin{equation}
(D_2)_{ii} = \sum_{k=1}^n [\psi(y_k)]_i [\psi(y_k)]_i
\end{equation}

\end{itemize}

(Singular value decomposition step)

Calculate $m$-rank thin SVD on $D_1^{-\frac{1}{2}} \Omega D_2^{-\frac{1}{2}}$.
Let $U$ and $V$ be the two resulting projection matrices. Return the two
functions $u(x) = (D_1^{-\frac{1}{2}}U)^{\top} \phi(x)$ and $v(y) = (D_2^{-\frac{1}{2}}V)^{\top} \psi(y)$.

}}
}
\caption{The CCA learning algorithm.\label{figure:alga}}
\end{figure}

\begin{equation}
\decoder(x) = \arg\max_{y \in \outy} \rho(u(x), v(y)). \label{eq:d}
\end{equation}

Our key observation is that one can use canonical correlation analysis to learn the two projections $u$ and $v$.
This is similar to the observation made by \newcite{udupa2010transliteration} and \newcite{jagarlamudi2012regularized} in previous work about transliteration.
The learning algorithm assumes the existence of two feature functions $\phi \colon \inx \rightarrow \mathbb{R}^{d \times 1}$
and $\psi \colon \outy \rightarrow \mathbb{R}^{d' \times 1}$, where $d$ and $d'$ could potentially be large, and the feature
functions could potentially lead to sparse vectors.

We then apply a modified version of canonical correlation analysis on the two ``views:'' one view corresponds to the input
feature function and the other view corresponds to the output feature function. This means we calculate the following
three matrices $D_1 \in \mathbb{R}^{d \times d}, D_2 \in \mathbb{R}^{d' \times d'}$ and $\Omega \in \mathbb{R}^{d \times d'}$:

\begin{align}
D_1 & = \mathrm{diag}\left(\displaystyle\frac{1}{n} \sum_{i=1}^n \phi(x_i) (\phi(x_i))^{\top}\right) \\
D_2 & = \mathrm{diag}\left(\displaystyle\frac{1}{n} \sum_{i=1}^n \psi(y_i) (\psi(y_i))^{\top}\right) \\
\Omega & = \displaystyle\frac{1}{n} \sum_{i=1}^n \phi(x_i) (\psi(y_i))^{\top}
\end{align}

\noindent where $\mathrm{diag}(A)$ for a square matrix $A$ is a diagonal matrix with the diagonal
copied from $A$. We then apply thin singular value decomposition on $D_1^{-1/2} \Omega D_2^{-1/2}$ so that

\begin{align}
D_1^{-1/2} \Omega D_2^{-1/2} \approx U \Sigma V^{\top},
\end{align}

\noindent with $U \in \mathbb{R}^{d \times m}$, $\Sigma \in \mathbb{R}^{m \times m}$ is a diagonal matrix of singular values
and $V \in \mathbb{R}^{d' \times m}$. The value of $m$ should be relatively small compared to $d$ and $d'$.
We then choose $u$ and $v$ to be:

\begin{align}
u(x) & = (D_1^{-\frac{1}{2}}U)^{\top} \phi(x), \\
v(y) & = (D_2^{-\frac{1}{2}}V)^{\top} \psi(y).
\end{align}

\begin{figure*}

\begin{center}

\begin{tikzpicture}[
   scale = 1,
   angle radius = 7mm,
my angle/.style = {draw,
                   angle eccentricity=1.2,
                   font=\footnotesize,
                   <->}, 
    Arrow/.style= {ultra thick,gray,-stealth,
                   shorten <=7mm, 
                   draw=gray}
                    ]
\coordinate  (c) at (0,0);
\draw[thick] (c) circle(1.3cm);
\foreach \x [count=\i] in {30,45}
{
    \filldraw[gray] (\x:1.3cm) circle(1pt);
    \draw[black,->] (c) -- coordinate (m\i) (\x:1.3cm) coordinate (a\i);
}
\pic [my angle,"$\theta$",red,angle radius = 10mm,]  {angle = a1--c--a2};
\node (li)  [draw, minimum size=20mm,align=center,left]  at (-2,0) {\includegraphics[width=1.3in]{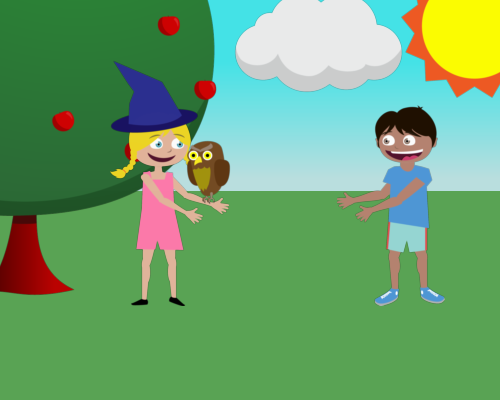}};
\node (ri)  [draw, minimum size=20mm,align=center,right] at ( 2,0) {Jenny is holding an owl.};
    \draw[Arrow]    (li.center) to [out=45, in=135]  (m2);
    \draw[Arrow]    (ri.center) to [out=225,in=-60]  (m1);
\end{tikzpicture}

\end{center}

\caption{Demonstration of CCA inference. An object from the input space $\inx$ (the image on the left $x$) is mapped to a unit
vector. Then, we find the closest unit vector which has an embodiment in the output space, $\outy$. That embodiment is the text on
the right, $y$. It also also holds that $\rho(u(x),v(y)) = \cos \theta$.\label{fig:cca-infer}}

\end{figure*}

For the similarity metric, we use the cosine similarity:

\begin{align}
\rho(z,z') & = \displaystyle\frac{\sum_{i=1}^m {z_i z'_i}}{\sqrt{\sum_{i=1}^m z_i^2} \sqrt{\sum_{i=1}^m (z'_i)^2}} \\
& = \displaystyle\frac{\langle z, z' \rangle}{||z||\cdot||z'||}.
\end{align}

Figure~\ref{fig:cca-infer} describes a sketch of our CCA inference algorithm.

\paragraph{Motivation} What is the motivation behind this use of CCA and the chosen projection matrices and similarity
metric? \newcite{osborne-16} showed that CCA maximizes the following objective:

\begin{equation}
\sum_{i,j} d_{ij} - n \sum_{i=1}^n d_{ii}^2,
\end{equation}

\noindent where

\begin{equation}
d_{ij} = \sqrt{\displaystyle\frac{1}{2} \left( \sum_{k=1}^m (u(x_i) - v(y_j))^2 \right)}.
\end{equation}

The objective is maximized with respect to the projections that CCA finds, $u$ and $v$.
This means that CCA finds projections such that the Euclidean distance between
$u(x)$ and $v(y)$ for matching $x$ and $y$ is minimized, while it is maximized for $x$ and $y$
that have a mismatch between them.

As such, it is well-motivated to use a similarity metric $\rho(u(x),v(y))$ which is inversely monotone with
respect to the Euclidean distance between $u(x)$ and $v(y)$.  
We next note that for any two vectors $z$ (denoting $u(x)$) and $z'$ (denoting $v(y)$) it holds
that (by simple algebraic manipulation):

\begin{align} 
- \langle z, z' \rangle & = \frac{1}{2} \left( || z - z'||^2 - || z||^2 - ||z'||^2 \right). \label{eq:aa}
\end{align}

This means that if the norms of $z$ and $z'$ are constant, then maximizing the cosine similarity
between $z$ and $z'$ is the same as minimizing Euclidean distance between $z$ and $z'$.
In our case, the norms of $u(x)$ and $v(y)$ are not constant, but we find that our algorithm is
much more stable when the cosine similarity is used instead of the Euclidean distance.

We also note that in order to minimize the distance between $z$ and $z'$ to follow CCA, according
to Eq.~\refeq{eq:aa}, we need to {\em maximize} the dot product between $z$ and $z'$ while minimizing
the norm of $z$ and $z'$. This is indeed the recipe that the cosine similarity metric follows.

In Section~\ref{section:decoding} we give an additional interpretation to the use of cosine similarity,
as finding the maximum aposteriori solution for a re-normalized von Mises-Fisher distribution.

\subsection{When the Output Space is Language}
\label{section:decoding}

While our approach to mapping from an input space to an output space through CCA is rather abstract and
general, in this paper we focus in cases where the output space $\outy \subseteq \Lambda^{\ast}$ is a set of strings over
some alphabet $\Lambda$, for example, the English language. For example, $\outy$ could be the set of all $n$-gram chains
possible over some $n$-gram set or the set of possible composition of atomic phrases, similar to phrase tables
in phrase-based machine translation \cite{koehn2007moses}.

For the language-vision problem we address in Section~\ref{section:experiments}, we assume the existence of a phrase table $\phrasetable$,
such that every $y \in \outy$ can be decomposed into a sequence of consecutive phrases $p_1,\ldots,p_r \in \phrasetable$.

\begin{figure}[t!]
{\small
\framebox{\parbox{3in}{
{\bf Inputs:} An input example $x$, a similarity metric $\rho$, two projection functions $u$ and $v$, a probabilistic phrase table $Q$, a constant $\eta \ge 0$,
a constant $\tau \in (0,1)$, a starting temperature $T$.

{\bf Algorithm:}

(Initialization)
\begin{itemize}

\item Let $y^{\ast}$ be an arbitrary point in the output space.

\item Let $y'$ be $y^{\ast}$.

\item Let $t$ be $T$.

\end{itemize}

While the temperature $t$ goes below a given value:

\begin{itemize}

\item Choose uniformly two different integers between $1$ and $|y'|$, the length of $y'$, $i$ and $j$.

\item Choose randomly a phrase $p$ from $Q(p \mid y'_{i-1}, y'_{j+1})$.

\item Let $y = y'_1 \cdots y'_{i-1} p y'_{j+1} \cdots y'_{|y'|}$.

\item If $\rho(u(x),v(y)) + \eta |y| \ge \rho(u(x),v(y^{\ast})) + \eta |y'|$, then set $y^{\ast}$ to be $y$.

\item Let

\begin{align}
\alpha_0 & = \frac{\exp\left(\displaystyle\frac{1}{t} \rho(u(x),v(y)) + \eta |y|\right)}{\exp\left(\displaystyle \frac{1}{t} \rho(u(x),v(y')) + \eta |y'| \right)} \\
\alpha_1 & = \frac{|y|^2 Q( y'_i\cdots y'_j \mid y'_{i-1},y'_{j+1})}{|y'|^2 Q( y_i\cdots y_j \mid y_{i-1},y_{j+1})} \\
\alpha & = \{ 1, \alpha_0 \cdot \alpha_1   \}.
\end{align}

\item Uniformly sample a number from $[0,1]$, and if it is smaller than $\alpha$, set $y'$ to be $y$.

\item Let $t \leftarrow \tau t$.

\end{itemize}

Return $y^{\ast}$.

}}
}
\caption{The CCA decoding algorithm.\label{figure:algb}}
\end{figure}

The problem of decoding over this space is not trivial. This is regardless of $\inx$ -- once $x$ is given, it is mapped
using $u(x)$ to a vector in $\mathbb{R}^m$, and at this point this is the information we use to further decode into $y$ -- 
the structure of $\inx$ before this transformation does not change much the complexity of the problem.
We propose the following approximate decoding algorithm $\decoder(x)$ for Eq.~\refeq{eq:d}. The algorithm is a Metropolis-Hastings (MH) algorithm that assumes
the existence of a blackbox sampler $q(y \mid y')$ -- the proposal distribution. This blackbox sampler randomly chooses two endpoints $k$ and $\ell$ in $y'$ and if possible,
replaces all the words in between these two words ($y'_k \cdots y'_{\ell}$) with a phrase $p \in \phrasetable$ such that in the training data, there is an occurrence of the new phrase
$p$ after the word $y'_{k-1}$ and before the word $y'_{\ell+1}$.

As such, we are required to create a probabilistic table of the form $Q \colon \Lambda \times \phrasetable \times \Lambda \rightarrow \mathbb{R}$ that
maps a pair of words $y,y' \in \Lambda$ and a phrase $p \in \phrasetable$ to the probability $Q(p \mid y,y')$. In our experiments, we use the phrase
table used by \newcite{ortiz2015learning}, extracted using Moses, and use relative frequency count to estimate $Q$: we count the number of times each
phrase $p$ appears between the context words $y$ and $y'$ and normalize.						

Since we are interested in {\em maximizing} the cosine similarity between $v(y)$ and $u(x)$, after each sampling step, we check whether
the cosine similarity of the new $y$ is higher (regardless of whether it is being accepted or rejected by the MH algorithm) than that of any $y$ so far.
We return the best $y$ sampled.

The ``true'' unnormalized distribution we use in the accept-rejection step is the exponentiated value of the cosine similarity between $u(x)$ and $v(y)$.
This means that for a given $x$, the MH algorithm implicitly samples from the following distribution $P$:

\begin{equation}
P(y \mid x) = \frac{\exp\left(\displaystyle\frac{\langle u(x), v(y) \rangle}{||u(x)|| \cdot || v(y)||}\right)}{Z(x)} \label{eq:a}
\end{equation}

\noindent where

\begin{equation}
Z(x) = \sum_{y' \in \outy} \exp\left(\displaystyle\frac{\langle u(x), v(y') \rangle}{||u(x)|| \cdot || v(y')||}\right).
\end{equation}

The probability distribution $P$ has a strong relationship to the von Mises-Fisher distribution, which is defined over vectors of unit vector.
The von Mises-Fisher distribution has a parametric density function $f(z ; \mu)$ which is proportional to the exponentiated dot product between the unit vector
$z$ and some other unit vector $\mu$ which serves as the parameter for the distribution. The main difference between the von Mises-Fisher distribution
and the distribution defined in Eq.~\refeq{eq:a} is that we do not allow {\em any} unit vector to be used as $\displaystyle\frac{v(y)}{||v(y)||}$ -- only
those which originate in some output structure $y$. As such, the distribution in Eq.~\refeq{eq:a} is a re-normalized version of the von-Mises distribution,
after elements from its support are removed.

In a set of preliminary experiments, we found that while our algorithm gives adequate descriptions to the images, it is not unusual for it to give short descriptions
that just mention a single object in the image. This relates to the adequacy-fluency tension that exists in machine translation problems. To overcome this issue, we
add to the cosine similarity a term $\eta |y|$ where $\eta$ is some positive constant tuned on a development set and $|y|$ is the
length of the sampled sentence. This pushes the decoding algorithm to prefer textual descriptions which are longer.

\paragraph{Simulated Annealing} Since we are not interested in {\em sampling} from the distribution $P(y \mid x)$, but actually find its mode,
we use simulated annealing with our MH sampler. This means that we exponentiate by a $\displaystyle\frac{1}{t}$ term the unnormalized distribution
we sample from, and decrease this temperature $t$ as the sampler advances. We start with a temperature $T= 10,000$, and multiply $t$ by $\tau = 0.995$ at each step. The idea is for the sampler to start with an exploratory phase, where it is jumping from different parts of the search space
to others. As the temperature decreases, the sampler makes smaller jumps with the hope that it has gotten closer to parts of the search space
where most of the probability mass is concentrated.

\section{Experiments}
\label{section:experiments}
\begin{table}
{\small
\begin{tabular}{|ccc|c|}
\hline
$y$ & $p$ & $y'$ & prob. \\
\hline
waiting & to & get & 1.000 \\
with & the & bucket. & 0.750 \\
pizza & on & the & 0.343 \\
trying & to get away from & jenny & 0.050 \\
baseball & with & the & 0.033 \\
is & playing near the & swings. & 0.011 \\
$\langle \mathrm{begin} \rangle$ & jenny is playing with a & colorful & 0.008 \\
is & surprised by the & owl & 0.006 \\
mike & and the bear are & standing & 0.002 \\
\hline
\end{tabular}
}
\caption{Example of phrases and their probabilities learned for the function $Q(p \mid y, y')$. The marker $\langle \mathrm{begin} \rangle$ marks the beginning of a sentence. \label{table:phrases}}
\end{table}

Our experiments are performed on a language-vision dataset, with the goal of taking a so-called ``abstract scene'' \cite{zitnick2013bringing}
and finding a suitable textual description. Figure~\ref{fig:cca-infer} gives a description of our CCA algorithm
in the context of this problem.

The Abstract Scenes Dataset consists of 10,020 scenes, each represented as a set of clipart objects placed in different
positions and sizes in a background image (consisting of a grassy area and sky).
Cliparts can appear in different ways, for example, the boy and the girl (cliparts 18 and 19), can be depicted 
sad, angry, sitting or running. The descriptions were given using crowdsourcing.

We use the same data split as \newcite{ortiz2015learning}. We use 7,014 of the scenes as a training set, 1,002 as a development
set and 2,004 as a test set.\footnote{Our dataset splits and other information can be found in \url{http://cohort.inf.ed.ac.uk/canonical-correlation-inference.html}.}
Each scene is labeled with at most eight short captions. We use all of these captions in the training set, leading to a total
of 42,276 training instances. We also use these captions as reference captions for both the development and the test set.


The feature function $\phi(x)$ for an image is based on the ``visual features'' that come with the abstract scene dataset.
More specifically, there are binary features that fire for 11 object categories, 58 specific objects,
co-occurrence of object category pairs,
co-occurrence of object instance pairs, absolute location of object categories and instances, absolute depth, relative location of objects,
relative location with directionality the object is facing, a feature indicating whether an object is near a child's hand or a child's head and attributes
of the children (pose and facial expression). The total number of features for this $\phi$ function is 7,149. See more in the description of the abstract scene dataset. 

The feature function $\psi(y)$ for an image description is defined as a one-hot representation for all phrases from the phrase
table of \newcite{ortiz2015learning} that fire in the image (the phrase table is denoted by $\phrasetable$ in Section~\ref{section:decoding}). This phrase table was obtained through the Moses toolkit \cite{koehn2007moses}.
The total number of phrases in this phrase table is 30,911. The size of the domain of $Q$ (meaning, the size of the phrase table with context words) is 120,019.
Table~\ref{table:phrases} gives a few example phrases and their corresponding probabilities.

In our CCA learning algorithm, we also need to decide on the value of $m$. We varied $m$ between $30$ and $300$ (in steps of $10$) and tuned its value on
the development set by maximizing BLEU score against the set of references.\footnote{We use the {\tt multeval} package from \url{https://github.com/jhclark/multeval}.} Interestingly enough, the BLEU scores did not change that much (they usually were within one point of each other for sufficiently large $m$), pointing to a stability of the algorithm with respect to the number of dimensions used.

\begin{figure}
\includegraphics[width=3.2in]{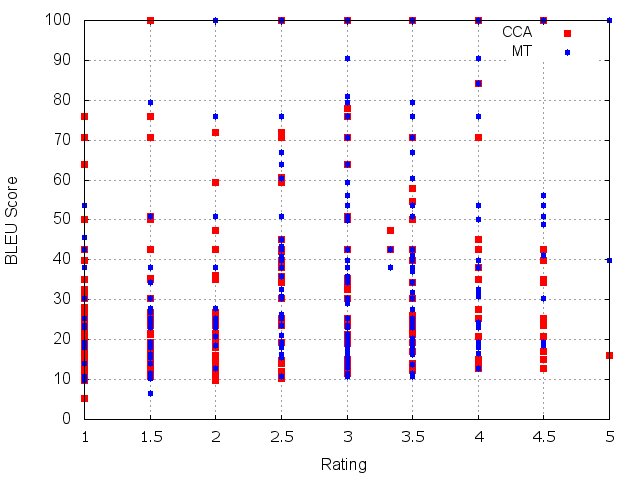}
\caption{Scatter plot of SMT (statistical machine translation) and CCA BLEU scores versus human ratings.\label{fig:scatter-plot}}
\end{figure}

\newcite{ortiz2015learning} partially measure the success of their system by comparing BLEU and METEOR scores of their different systems while using the descriptions given in the dataset as
a reference set. The scores for their different systems are given in Table~\ref{table:results1}. They compare their system (SMT, based on phrase-based machine translation) against several
baselines:\footnote{We note that we also experimented with a neural network model (\textsc{Seq2Seq} model), but it performed badly, giving a BLEU score of 10.20 and a METEOR score of 15.20 with inappropriate captions. It seems
like \textsc{Seq2Seq} models is unfit for this dataset, perhaps because of its size. See also \newcite{rastogi2016weighting} for similar results.}
\begin{itemize}
 \item \textbf{LBL}: a log-bilinear language model trained on the image captions only.
 \item \textbf{MLBL}: mutlimodal log-bilinear model, implementation of \newcite{kiros2014multimodal}.
 \item \textbf{Image}: a system that for every new image, queries the set of training images for the most similar one, and returns a random description of that training example. 
 \item \textbf{Keyword}: system that annotates every image with keywords that most probably describe it and then do a search query against all training data descriptions,
 returning the description that is closest (in terms of TF-IDF similarity) to the keywords.
 \item \textbf{Template}: system that uses templates inferred from dependency parses of the training data descriptions. A set of templates is discovered and
 a classifier that associates images with templates is trained.
 \item \textbf{SMT}: Ortiz et al. system first selects pairs of clipart objects that are important enough to be described by solving an integer linear programming problem, creates a ``visual
 encoding'' using visual dependency grammar and finally uses a phrase-based SMT engine to translate the latter to proper sentences.
\end{itemize}

Our system does not score as high as their machine translation system.

\begin{table}

\begin{center}
\begin{tabular}{|l|l|c|c|}
\hline
& system & BLEU & METEOR \\
\hline
\multirow{6}{*}{\rotatebox[origin=c]{90}{Ortiz et al.}} & LBL & 7.3 & 17.7 \\
& MLBL & 12.3 & 20.4 \\
& Image & 12.8 & 21.7 \\
& Keyword & 14.7 & 26.6 \\
& Template & 40.3 & 30.4 \\
& SMT & 43.7 & 35.6 \\
\hhline{~---}
& CCA & 26.1 & 25.6 \\
\hline
\end{tabular}
\end{center}

\caption{Scene description evaluation results on the test set, comparing the systems from Ortiz et al. to our CCA inference algorithm
(the first six results are reported from the Ortiz et al. paper). The CCA result uses $m=120$ and $\eta = 0.05$, tuned on the development set.
See text for details about each of the first six baselines.
\label{table:results1}}

\end{table}

\begin{figure}[t]

\begin{center}
\includegraphics[width=1.2in]{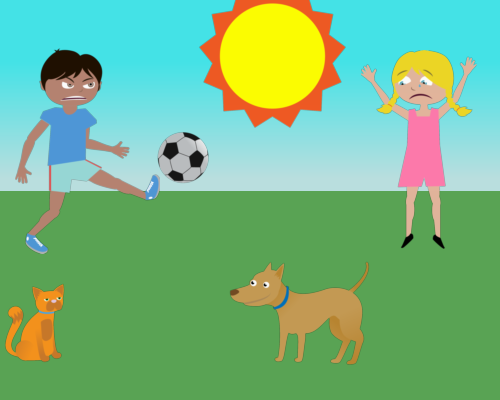}
\end{center}

\caption{An image with the following descriptions in the dataset:
(1) {\tt mike is kicking the soccer ball};
(2) {\tt mike is sitting on the cat};
(3) {\tt jenny is standing next to the dog};
(4) {\tt jenny is kicking the soccer ball};
(5) {\tt the sun is behind jenny};
(6) {\tt the soccer ball is under the sun}.
\label{fig:desc}}

\end{figure}

\begin{figure*}[t]
\begin{center}
\begin{tabular}{llll}
& \includegraphics[width=1.1in]{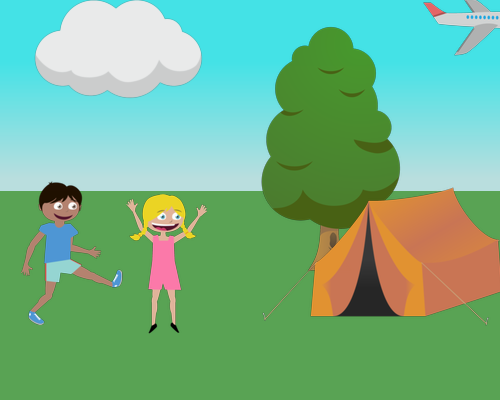} &
\includegraphics[width=1.1in]{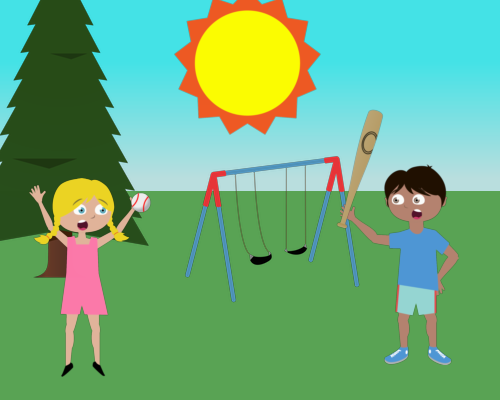} &
\includegraphics[width=1.1in]{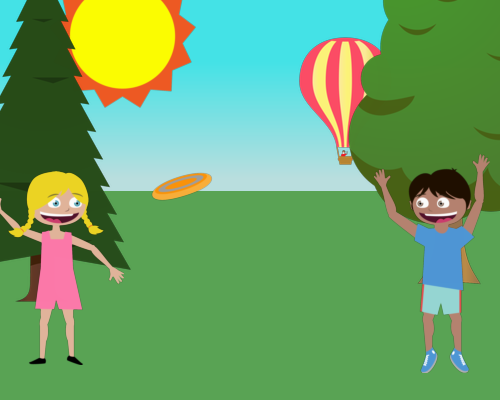} \\
SMT & jenny is waving at mike & jenny is wearing a baseball & jenny is holding a frisbee \\
CCA & mike and jenny are camping & mike is holding a bat & jenny is throwing the frisbee  \\
\\
& \includegraphics[width=1.1in]{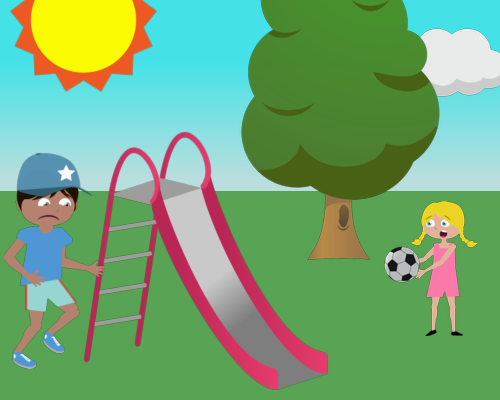} &
\includegraphics[width=1.1in]{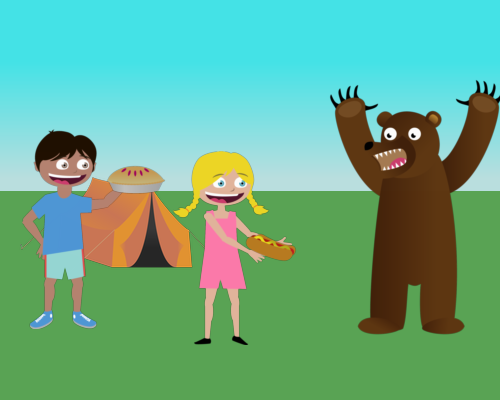} &
\includegraphics[width=1.1in]{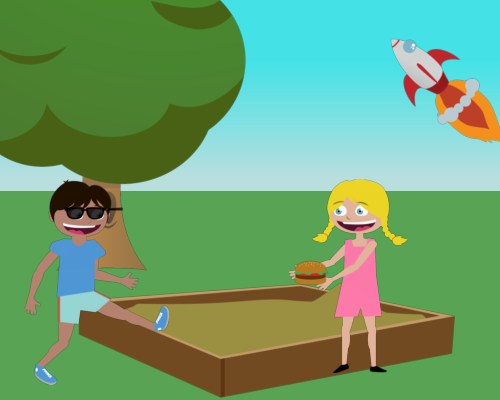} \\
SMT & jenny is kicking the soccer ball & jenny is holding a hot dog & jenny is holding a hamburger \\
CCA & mike is kicking a blass & jenny wants the bear & the rocket is behind mike
\end{tabular}
\end{center}
\caption{Examples of outputs from the machine translation system and from CCA inference. The top three images give examples where the CCA inference outputs were rated
highly by human evaluations (4 or 5), and the SMT ones were rated poorly (1 or 2). The bottom three pictures give the reverse case.\label{fig:examples}}
\end{figure*}

It is important to note that the descriptions given in the dataset, as well as those generated by the different systems are not ``complete.'' Each one of them describes
a specific bit of information that is implied by the scene. Figure~\ref{fig:desc} demonstrates this. As such, the calculation of SMT evaluation scores with respect to
a reference set is not necessarily the best mechanism to identify the correctness of a textual description.
To demonstrate this point, we measure BLEU scores of one of the reference
sentences while comparing it to the other references in the set. We did that for each of the eight batches of references available in the training set.

The average reference BLEU score is 24.1 and the average METEOR score is 20.0, a significant drop compared to the machine translation system of \newcite{ortiz2015learning}.
We concluded from this result that the SMT system is not ``creatively'' mapping the images to their corresponding descriptions. It relies heavily on the training set captions,
and learns how to map images to sentences in a manner which does not generalize very well outside of the training set.

\begin{table}
\begin{center}
\begin{tabular}{|c||c|c|}
\hline
slice & CCA $< 3$ & CCA $ \ge 3$ \\
\hline
\hline
SMT $< 3$ & M:1.77 & M:1.92 \\
         & C:1.64 & C:3.71 \\
\hline
SMT $\ge 3$ & M:3.42 & M:3.46 \\
         & C:1.47 & C:3.54 \\
\hline
\end{tabular}
\end{center}
\caption{Average ranking by human judges for cases in which the caption has an average rank of $3$ or higher (for both CCA and SMT) and when its average rank is lower than $3$. ``M'' stands for SMT average rank and ``C'' stands for CCA
average rank.\label{table:ranking}}
\end{table}

Another indication that our system creates a more diverse set of captions is that the number of {\em unique} captions it generates for the test set
is significantly larger than that of the SMT system by Ortiz et al. The SMT system generates 359 unique captions (out of 2,004 instances in the test set), while CCA
generates 496 captions, an increase of 38.1\%.

To test this hypothesis about caption diversity, we conducted the following human experiment. We asked 12 subjects to rate the captions of 300 abstract scenes with a score between 1 to
5.\footnote{The ratings can be found here: \url{http://cohort.inf.ed.ac.uk/canonical-correlation-inference.html}.} Each rater was presented with
three captions: a reference caption (selected randomly from the gold-standard captions), an SMT caption and a caption from our system (presented in a random order) 
and was asked to rate the captions on adequacy level (on a scale of 1 to 5). Most images
were rated exactly twice, with a few images getting three raters.
A score of 1 or 2 means that the caption likely does not adequately describe the scene. A score of 3 usually means
that the caption describes some salient component in the scene, but perhaps not the most important one. Scores of 4 and 5 usually denote good captions that adequately describe the corresponding scenes.
This experiment is similar to the one done by Ortiz et al.

The ranking results are given in Table~\ref{table:ranking}. The results show that our system tends to score higher for images which are highly ranked (by both the SMT system and CCA),
but tends to score lower for images which are lower ranked.

In addition, we checked the MT evaluation scores for highly ranked captions both for SMT and CCA (ranking larger than 4). For SMT, the BLEU scores are
49.70 (METEOR 40.10) and for CCA it is 41.80 (METEOR 33.10). This is not the result of images in SMT being ranked higher, as the average ranking
among these images is 4.18 for the SMT system and 4.25 for CCA. The lower CCA score again indicates that our system gives captions which are not necessarily aligned
with the references, but still correct. It also highlights the flaw with using MT evaluation metrics for this dataset. Figure~\ref{fig:scatter-plot} also demonstrates
that the correlation between BLEU scores and human ranking is not high. More specifically, in that plot, the correlation between the $x$-axis (ranking) and $y$-axis (BLEU
scores) for CCA is $0.3$ and for the SMT system $0.31$.



Figure~\ref{fig:examples} describes six examples in which the human raters rated the SMT system highly and CCA poorly and vice-versa.

\section{Conclusion}

We described a technique to predict structures from complex input spaces
to complex output spaces based on canonical correlation analysis.
Our approach projects the input space into a low-dimensional representation,
and then converts it back into an instance in the output space.
We demonstrated the use of our method on the structured prediction problem
of attaching textual captions to abstract scenes. Human evaluation of these
captions demonstrate that our approach is promising for generating text
from images.

\section*{Acknowledgments}

The authors would like to thank Mirella Lapata, Luis Mateos and Clemens Wolff
for their help with replicating the results from their paper.
Thanks also to Marco Damonte for useful comments.
This research was supported by the H2020 project SUMMA, under grant agreement 688139.

\bibliography{nlp}
\bibliographystyle{aaai}

\end{document}